\documentclass[runningheads]{llncs}

\usepackage{eccv}

\usepackage{eccvabbrv}

\usepackage{graphicx}
\usepackage{booktabs}

\usepackage[accsupp]{axessibility}  % Improves PDF readability for those with disabilities.
\RequirePackage{makecell}
\usepackage{xcolor,colortbl}
\usepackage{bbding}
\usepackage{color}
\usepackage{multirow}
\usepackage{multicol}

\RequirePackage{makecell}
\usepackage{xcolor,colortbl}
\usepackage{bbding}
\usepackage{color}
\usepackage{multirow}
\usepackage{multicol}
\usepackage{amssymb}
\usepackage{lipsum}
\usepackage{fontawesome}
\usepackage{academicons}
\usepackage[pagebackref,breaklinks,colorlinks,citecolor=eccvblue]{hyperref}
\usepackage{orcidlink}

\begin{document}

% ---------------------------------------------------------------
% TODO REVIEW: Replace with your title
\title{
MapDistill: Boosting Efficient Camera-based HD Map Construction via Camera-LiDAR Fusion Model Distillation
}

% TODO REVIEW: If the paper title is too long for the running head, you can set
% an abbreviated paper title here. If not, comment out.
\titlerunning{MapDistill: HD Map Construction via Model Distillation}

% TODO FINAL: Replace with your author list. 
% Include the authors' OCRID for the camera-ready version, if at all possible.
\author{Xiaoshuai Hao$^{1}$$^{\dagger}$ \and Ruikai Li$^{2}$$^{\dagger}$ \and Hui Zhang$^{1}$ \and Dingzhe Li$^{1}$ \and Rong Yin$^{3}$ \and Sangil Jung$^{4}$ \and Seung-In Park$^{4}$ \and  ByungIn Yoo$^{4}$ \and Haimei Zhao$^{5}$$^{\S}$ \and Jing Zhang$^{5}$$^{\S}$}

% TODO FINAL: Replace with an abbreviated list of authors.
\authorrunning{Xiaoshuai Hao et al.}
% First names are abbreviated in the running head.
% If there are more than two authors, 'et al.' is used.

% TODO FINAL: Replace with your institution list.
%\vspace{1em}

\institute{$^{1}$~Samsung R\&D Institute China–Beijing 
\footnotetext{$\dagger$~The first two authors contributed equally to this work.}
\footnotetext{$\S$~Corresponding authors.}\\
\quad
$^{2}$~State Key Lab of Intelligent Transportation System, Beihang University\\
 \quad
$^{3}$~Institute of Information Engineering, Chinese Academy of Sciences\\
 \quad
$^{4}$~Computer Vision TU, SAIT, SEC, Korea
\quad
$^{5}$~The University of Sydney\\
\email{\{xshuai.hao,
hui123.zhang,
dingzhe.li, byungin.yoo\}@samsung.com} \\ ricky$\_$developer@buaa.edu.cn \quad \{si14.park, sang-il.jung\}@samsung.com \\
yinrong@iie.ac.cn \quad hzha7798@uni.sydney.edu.au \quad  jingzhang.cv@gmail.com}

\maketitle

\begin{abstract}
Online high-definition (HD) map construction is an important and challenging task in autonomous driving. Recently, there has been a growing interest in cost-effective multi-view camera-based methods without relying on other sensors like LiDAR. However, these methods suffer from a lack of explicit depth information, necessitating the use of large models to achieve satisfactory performance. To address this, we employ the Knowledge Distillation (KD) idea for efficient HD map construction for the first time and introduce a novel KD-based approach called MapDistill to transfer knowledge from a high-performance camera-LiDAR fusion model to a lightweight camera-only model. Specifically, we adopt the teacher-student architecture, \ie, a camera-LiDAR fusion model as the teacher and a lightweight camera model as the student, and devise a dual BEV transform module to facilitate cross-modal knowledge distillation while maintaining cost-effective camera-only deployment. Additionally, we present a comprehensive distillation scheme encompassing cross-modal relation distillation, dual-level feature distillation, and map head distillation. This approach alleviates knowledge transfer challenges between modalities, enabling the student model to learn improved feature representations for HD map construction. Experimental results on the challenging nuScenes dataset demonstrate the effectiveness of MapDistill, surpassing existing competitors by over 7.7 mAP or 4.5$\times$ speedup. 
%The source code will be released.

\keywords{HD Map Construction \and Knowledge Distillation \and Lightweight }
\end{abstract}

\section{Introduction}
\label{sec:intro}

Online high-definition (HD) map provides abundant and precise static environmental information about the driving scenes, which is fundamental for planning and navigation in autonomous driving systems.
Recently, multi-view camera-based~\cite{MapTR,ding2023pivotnet,qiao2023end} HD map construction has gained increasing attention thanks to the significant progress of Bird’s-Eye-View (BEV) perception. 
Compared with LiDAR-based~\cite{wang2023lidar2map,2020fisharixv} and Fusion-based methods \cite{li2022hdmapnet,liu2023vectormapnet,MapTR}, multi-view camera-based methods can be deployed at low cost, while the lack of depth information makes current approaches adopt large models for effective feature extraction and good performance achievement.
Therefore, it is crucial to trade off the performance and efficiency of the camera-based model for practical deployment.

\begin{figure}[t]
	\setlength{\abovecaptionskip}{-0.1cm}
	\begin{center}
		\includegraphics[width=0.83\linewidth]{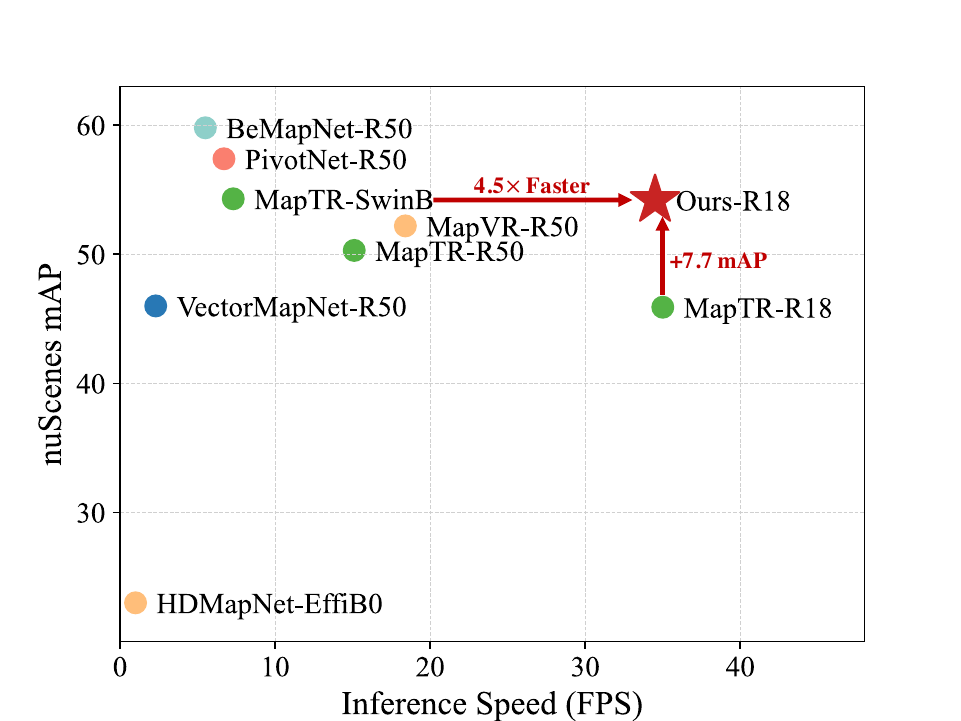}
	\end{center}
	\caption{
       Comparison of different methods on the nuScenes val dataset.
       We benchmark the inference speed on a single NVIDIA RTX 3090 GPU.
    Our method can achieve a better trade-off in both speed (FPS) and accuracy (mAP).
	}
	\vspace{-2.5em}
	\label{fig1}
\end{figure}

To achieve this goal, Knowledge Distillation (KD)~\cite{gou2021knowledge} has drawn great attention in related fields since it is one of the most practical techniques for training efficient yet accurate models.
KD-based methods usually transfer knowledge from a large well-trained model (teacher) to a small model (student) ~\cite{hinton2015distilling}, which has made remarkable progress in many fields, such as image classification~\cite{mirzadeh2020improved}, 2D object detection~\cite{chen2017learning}, semantic segmentation~\cite{yan20222dpass} and 3D object detection~\cite{chen2022bevdistill,zhou2023unidistill,zhao2023bevsimdet}.
Previous methods follow the well-known teacher-student paradigm~\cite{hinton2015distilling}, which forces the logits of the student network to match those of the teacher network.
Recently, BEV-based KD methods have advanced the field of 3D object detection, which unify the image and LiDAR features in the Bird-Eye-View (BEV) space and adaptively transfer knowledge across non-homogenous representations in a teacher-student paradigm.
Existing works use a strong LiDAR teacher model to distill a camera student model, such as BEVDistill~\cite{chen2022bevdistill}, UVTR~\cite{li2022unifying}, BEVLGKD~\cite{li2022bev}, TiG-BEV~\cite{huang2022tig} and DistillBEV~\cite{wang2023distillbev}.
Furthermore, the latest work UniDistill~\cite{zhou2023unidistill} proposes a universal cross-modality knowledge distillation framework for 3D Object Detection.

Compared to these methods, BEV-based HD map construction KD method differs in two crucial aspects: 
Firstly, the detection head (DetHead) produces the output of classification and localization for objects, while the output of the map head (MapHead) from a vectorized map construction model, \eg MapTR~\cite{MapTR}, is the classification and point regression result.
Secondly, existing BEV-based KD methods for 3D object detection typically focus on aligning foreground objects' features to mitigate the background environment's adverse impact, which is obviously unsuitable for HD map construction.
Therefore, directly applying the BEV-based KD method for 3D object detection to HD map construction fails to achieve satisfying results (see the experiment results in Tab.~\ref{tab1}) due to the inherent dissimilarity between the two tasks. 
To the best of our knowledge, BEV-based KD methods for HD map construction are still under exploration.

To fill this gap, we propose a novel KD-based method named MapDistill to transfer the knowledge from a high-performance teacher model to an efficient student model. First, we adopt the teacher-student architecture, \textit{i.e.}, a camera-LiDAR fusion model as the teacher and a lightweight camera model as the student, and devise a dual BEV transform module to facilitate cross-modal knowledge distillation while maintaining cost-effective camera-only deployment. 
Building upon this architecture, we propose a comprehensive distillation scheme encompassing cross-modal relation distillation, dual-level feature distillation, and map head distillation, to mitigate the knowledge transfer challenges between modalities and help the student model learn improved feature representations for HD map construction. 
Specifically, we first introduce the cross-modal relation distillation loss for the student model to learn better cross-modal representations from the fusion teacher model.
Second, to achieve better semantic knowledge transfer, we employ the dual-level feature distillation loss on both the low-level and high-level feature representations in the unified BEV space.
Last but not least, we specifically introduce a map head distillation loss tailored for the HD map construction task, including classification loss and point-to-point loss, which can make the final predictions of the student closely resemble those of the teacher.
Extensive experiments on the challenging nuScenes dataset~\cite{20cvprnuscense} demonstrate the effectiveness of MapDistill, surpassing existing competitors by over 7.7 mAP or 4.5$\times$ speedup as shown in Fig.~\ref{fig1}.

The contributions of this paper are mainly three-fold:
\begin{itemize}
\item 
We present an effective model architecture for distillation-based HD map construction, including a camera-LiDAR fusion teacher model, a lightweight camera-only student model, and a dual BEV transform module, which facilitates knowledge transfer within and between different modalities while enjoying cost-effective camera-only deployment.
\item 
We introduce a comprehensive distillation scheme that supports cross-modal relation distillation, dual-level feature distillation, and map head distillation simultaneously. By mitigating the knowledge transfer challenges between modalities, this distillation scheme helps the student model learn better feature representation for HD map construction.
\item

MapDistill achieves superior performance than state-of-the-art (SOTA) methods, which could serve as a strong baseline for KD-based HD map construction research.
\end{itemize}

\section{Related Work}
\label{sec:related}

\textbf{Camera-based HD Map Construction.}
HD map construction is a prominent and extensively researched area within the field of autonomous driving.
Recently, camera-based methods~\cite{zhang2023online,ding2023pivotnet,qiao2023end,li2022hdmapnet,liu2023vectormapnet,MapTR,hao2024your,hao2024team,kong2024robodrive} 
have increasingly employed the Bird's-eye view (BEV) representation as an ideal feature space for multi-view perception due to its remarkable ability to mitigate scale-ambiguity and occlusion challenges.
%%%%%%%%%%
Various techniques have been proposed and utilized to project perspective view (PV) features into the BEV space by leveraging geometric priors, such as 
LSS~\cite{philion2020lift}, Deformable Attention~\cite{22eccvbevformer} and GKT~\cite{2022GKT}.
%%%%%%%%%%
Furthermore, camera-based methods have come to rely on higher resolution images and larger backbone models to achieve enhanced accuracy~\cite{liu2021Swin,liu2021swinv2,wang2022internimage,22eccvbevformer,yang2022bevformer,xiong2023neural,22arxivbeverse,li2024foundation}, a practice that introduces substantial challenges for practical deployment.
For example, HDMapNet~\cite{li2022hdmapnet} and VectorMapnet~\cite{liu2023vectormapnet} employ the Efficient-B0 model~\cite{tan2019efficientnet} and ResNet50 model, respectively, as backbones for feature extraction.
Additionally, MapTR~\cite{MapTR} explores the impact of various backbones, including the Swin Transformer~\cite{liu2021Swin}, ResNet50~\cite{2016cvprresnet}, and Resnet18~\cite{2016cvprresnet}. 
Experimental results demonstrate a direct correlation between the backbone's representation capability and model performance, \ie, larger models generally yield better results. Yet, using larger models leads to slower inference, compromising the cost advantage of camera-based methods.
In this paper, we introduce an effective yet efficient camera-based method tailored for practical deployment via knowledge distillation.

\textbf{Fusion-based HD Map Construction.}
LiDAR-based methods~\cite{wang2023lidar2map,li2022hdmapnet,liu2023vectormapnet,2020fisharixv,haoMBFusion} provide precise spatial data for creating the BEV feature representation.  
Recently, camera-LiDAR fusion methods~\cite{li2022hdmapnet,MapTR,liu2023bevfusion,22icarv,tang2023thma,liang2022bevfusion,maptrv2,borse2023x} leverage the semantic richness of camera data and the geometric information from LiDAR in a collaborative manner. This fusion at the BEV level incorporates distinct streams, encoding camera and LiDAR inputs into shared BEV features, surpassing unimodal input approaches in performance. However, this integration may impose significant computational and cost burdens in practical deployment. To address this issue, we leverage KD techniques for efficient HD map construction and introduce a novel approach called MapDistill to transfer knowledge from a high-performance camera-LiDAR fusion model to a lightweight camera-only model, yielding a cost-effective yet accurate solution.

\textbf{Knowledge Distillation.}
KD refers to transferring
knowledge from a well-trained, larger teacher model to a smaller %, untrained 
student~\cite{hinton2015distilling}, which has been widely applied across diverse tasks, such as image classification~\cite{zhang2022quantifying,zhang2022dtfd,mirzadeh2020improved}, 2D object detection~\cite{zheng2023localization,chen2017learning}, semantic segmentation~\cite{yan20222dpass,yang2023label,shang2023incrementer,wang2023lidar2map} and 3D object detection~\cite{chen2022bevdistill,zhou2023unidistill,zhao2023bevsimdet,cho2023itkd}.
Recently, BEV-based KD methods have gained increasing attention in the field of 3D object detection.
Several existing works have adopted cross-modality knowledge distillation frameworks for 3D object detection, including BEVDistill~\cite{chen2022bevdistill}, UVTR~\cite{li2022unifying}, BEV-LGKD~\cite{li2022bev}, TiG-BEV~\cite{huang2022tig}, DistillBEV~\cite{wang2023distillbev}, and UniDistill~\cite{zhou2023unidistill}. Despite the numerous KD methods for 3D object detection, KD-based HD map construction remains relatively under-explored. 
In this paper, we fill this gap by proposing a novel KD-based approach called MapDistill to boost efficient camera-based HD map construction via camera-LiDAR fusion model distillation.

\begin{figure*}[t]
	\centering
 	\includegraphics[width=1.0\textwidth]{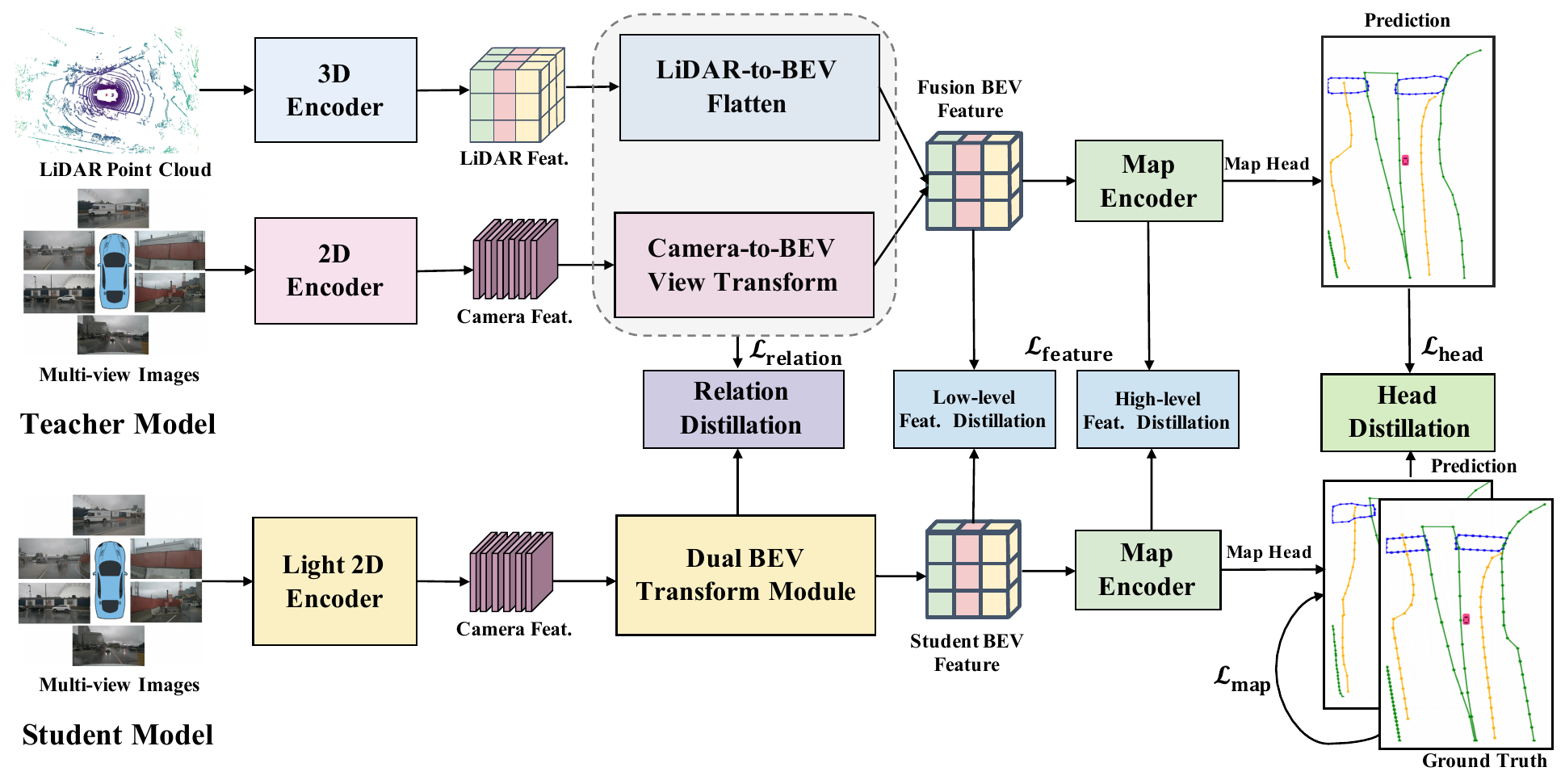}
	\caption{
\textbf{The overview of our proposed MapDistill.}
It consists of a fusion-based teacher model (top) and a lightweight
camera-based student model (bottom).
In addition, three distillation losses are employed to enable the teacher model to transfer knowledge to the student, \textit{i.e.}, by instructing the student model to produce similar features and predictions, which are cross-modal relation distillation ($\mathcal{L}_{relation}$), dual-level feature distillation ($\mathcal{L}_{feature}$), and map head distillation ($\mathcal{L}_{head}$). Note that only the student model is needed for  inference.
}
\vspace{-1em}
	\label{fig2}
\end{figure*}

\section{Methodology}
In this section, we describe our proposed MapDistill in detail.
We first give an overview of the whole framework in Fig.~\ref{fig2} and clarify the model designs of the teacher and student models in Sec.~\ref{sec3.1}.
Then, we elaborate details of MapDistill objectives in Sec.~\ref{sec3.2}, such as the cross-modal relation distillation, the dual-level feature distillation, and the map head distillation.
Finally, we present the overall training procedure %for our method 
in Sec.~\ref{sec3.3}.

\subsection{Model Overview}
\label{sec3.1}

\textbf{Fusion-based Model (Teacher).}
To enable the knowledge transfer from the camera-LiDAR fusion teacher model to the student model, we first establish a baseline of fusion-based HD map construction based on the state-of-the-art MapTR~\cite{MapTR} model. 
The fused MapTR model has two branches, as depicted in the top part of Fig.~\ref{fig2}.
For the camera branch, it firstly utilizes \textbf{Resnet50}~\cite{2016cvprresnet} as the backbone to extract multi-view features. 
Next, it uses GKT \cite{2022GKT} as the 2D-to-BEV transformation module to convert the multi-view features into the BEV space.
The generated camera BEV features can be denoted as $\textbf{F}_{C_{bev}}^{T} \in \mathbb{R}^{H\times W\times C}$, where $H, W, C$ represents the height, width and the number of channels of BEV features respectively, and the superscript $T$ is short for ``teacher''.
For the LiDAR branch, it adopts \textbf{SECOND}~\cite{2018seoncd} for point cloud voxelization and LiDAR feature encoding. 
The LiDAR features are projected to BEV space using a flattening operation as in \cite{liu2023bevfusion}, to obtain the LiDAR BEV representation $\textbf{F}_{L_{bev}}^{T} \in \mathbb{R}^{H\times W\times C}$.
Then, MapTR concatenates $\textbf{F}_{C_{bev}}^{T}$ and  $\textbf{F}_{L_{bev}}^{T}$ and processes the features with the fully convolutional network to produce the fused BEV features  $\textbf{F}_{fused}^{T} \in \mathbb{R}^{H\times W\times C}$.

The following step is to use a Map Encoder (MapEnc), which takes the fused BEV features $\textbf{F}_{fused}^{T}$ as input, to further generate the high-level feature $\textbf{F}_{high}^{T}$:
\begin{equation}
\label{eq1}
\textbf{F}_{high}^{T} = \operatorname{MapEnc}(\textbf{F}_{fused}^{T}),
\end{equation}
Then, the teacher Map head (MapHead) employs the classification and point branches to produce the final predictions of map elements categories $\textbf{F}_{cls}^{T}$ and point positions $\textbf{F}_{point}^{T}$:
\begin{equation}
\label{eq2}
\textbf{F}_{cls}^{T},\textbf{F}_{point}^{T} = \operatorname{MapHead}(\textbf{F}_{high}^{T}).
\end{equation}
During the overall training procedure, the teacher model will continuously produce diverse features $\textbf{F}_{C_{bev}}^{T}$, $\textbf{F}_{L_{bev}}^{T}$, $\textbf{F}_{fused}^{T}$, $\textbf{F}_{high}^{T}$, $\textbf{F}_{cls}^{T}$ and $\textbf{F}_{point}^{T}$.

\textbf{Camera-based Model (Student).}
To realize real-time inference speed for practical deployment, we adopt MapTR's camera branch as the base for the student model. 
Note that we employ \textbf{Resnet18}~\cite{2016cvprresnet} as the backbone to extract the multi-view features, which can make the network lightweight and easy to deploy.
On the base from MapTR, to mimic the multimodal fusion pipeline of the teacher model, 
we propose a Dual BEV Transform module to convert the multi-view features into two distinct BEV subspaces, whose effect will be verified in the ablation experiments. 
Specifically, we firstly use GKT~\cite{2022GKT} to generate BEV features in the first subspace $\textbf{F}_{C_{sub1}}^{S} \in \mathbb{R}^{H\times W\times C}$, where the superscript $S$ is short for ``student''.
Then, we utilize LSS~\cite{philion2020lift} to generate BEV features in the second subspace $\textbf{F}_{C_{sub2}}^{S} \in \mathbb{R}^{H\times W\times C}$.
Then, we concatenate $\textbf{F}_{C_{sub1}}^{S}$ and  $\textbf{F}_{C_{sub2}}^{S}$ and process the features with the fully convolutional network to produce the fused BEV features  $\textbf{F}_{fused}^{S} \in \mathbb{R}^{H\times W\times C}$.

Then, employing the same process as the teacher model, we can generate  $\textbf{F}_{high}^{S}$, $\textbf{F}_{cls}^{S}$ and $\textbf{F}_{point}^{S}$ from $\textbf{F}_{fused}^{S}$ with Eq.~\ref{eq1} and Eq.~\ref{eq2}. 
Therefore, the student model will consistently produce $\textbf{F}_{C_{sub1}}^{S}$, $\textbf{F}_{C_{sub2}}^{S}$, $\textbf{F}_{fused}^{S}$, $\textbf{F}_{high}^{S}$, $\textbf{F}_{cls}^{S}$ and $\textbf{F}_{point}^{S}$ during the procedure of map construction.

\subsection{MapDistill Objectives}
\label{sec3.2}

\subsubsection{Cross-modal Relation Distillation.}
The teacher model, a camera-LiDAR fusion model, combines semantic-rich information from camera data with explicit geometric data from LiDAR. In contrast, the student model, a camera-based model, focuses mainly on capturing semantic information from the camera. The essential factor contributing to the teacher model's superior performance is cross-modal interaction, which the student model lacks. Therefore, we encourage the student model to develop this cross-modal interaction capability through imitation.

To this end, we introduce a cross-modal attention distillation objective.
The core idea is to let the student model imitate the cross-modal attention of the teacher model during training. More specifically, for the teacher model, we begin by reshaping the camera BEV features $\textbf{F}_{C_{bev}}^{T} \in \mathbb{R}^{H\times W\times C}$ and the LiDAR BEV features $\textbf{F}_{L_{bev}}^{T} \in \mathbb{R}^{H\times W\times C}$ into sequences of 2D patches represented as $\textbf{Fp}_{C_{bev}}^{T} \in \mathbb{R}^{N\times (P^2C)}$ and $\textbf{Fp}_{L_{bev}}^{T} \in \mathbb{R}^{N\times (P^2C)}$, respectively. Here, the patch size is denoted as $P\times P$, and the number of patches is given by $N = HW/P^2$.

%%%%%%%%%%%%%%%%%%%%%
Then, we calculate the cross-modal attention from the teacher, including camera-to-lidar attention $\textbf{A}_{c2l}^{T} \in \mathbb{R}^{N\times N}$ and lidar-to-camera attention $\textbf{A}_{l2c}^{T} \in \mathbb{R}^{N\times N}$ as follows:
\begin{equation}
\label{eq3}
\begin{array}{l}
A_{c 2 l}^{T}=\operatorname{softmax}\left(\frac{\textbf{Fp}_{C_{bev}}^{T} \operatorname{Transpose}\left(\textbf{Fp}_{L_{bev}}^{T}\right)}{\sqrt{D_{k}}}\right)
\end{array},
\end{equation}
\begin{equation}
\label{eq4}
\begin{array}{l}
A_{l 2 c}^{T}=\operatorname{softmax}\left(\frac{\textbf{Fp}_{L_{bev}}^{T} \operatorname{Transpose}\left(\textbf{Fp}_{C_{bev}}^{T}\right)}{\sqrt{D_{k}}}\right)
\end{array},
\end{equation}
%where $D_{k}$ denotes 
where $\frac{1}{\sqrt{D_k}}$ is a scaling factor for preventing the softmax function from falling into a region with extremely small gradients when the magnitude of dot products grow large.

For the student model, we adopt the same operation as the teacher model to generate $\textbf{Fp}_{C_{sub1}}^{S} \in \mathbb{R}^{N\times (P^2C)}$ and  $\textbf{Fp}_{C_{sub2}}^{S} \in \mathbb{R}^{N\times  (P^2C)}$ from $\textbf{F}_{C_{sub1}}^{S}$ and $\textbf{F}_{C_{sub2}}^{S}$, respectively, and then compute the cross-modal attention of the student $\textbf{A}_{c2l}^{S}$, $\textbf{A}_{l2c}^{S}$ as follows: %zhh: if not enough space, we may delete the two formulas below, just refer to the previous formula for the teacher. 
\begin{equation}
\label{eq3s}
\begin{array}{l}
A_{c 2 l}^{S}=\operatorname{softmax}\left(\frac{\textbf{Fp}_{C_{sub1}}^{S} \operatorname{Transpose}\left(\textbf{Fp}_{C_{sub2}}^{S}\right)}{\sqrt{D_{k}}}\right)
\end{array},
\end{equation}
\begin{equation}
\label{eq4s}
\begin{array}{l}
A_{l 2 c}^{S}=\operatorname{softmax}\left(\frac{\textbf{Fp}_{C_{sub2}}^{S} \operatorname{Transpose}\left(\textbf{Fp}_{C_{sub1}}^{S}\right)}{\sqrt{D_{k}}}\right)
\end{array}.
\end{equation}

\iffalse
%%%%%%%%%%%%%%%%%%%%%
Then, we calculates the teacher cross-modal attention distribution, including camera-to-lidar $\textbf{A}_{c2l}^{T} \in \mathbb{R}^{N\times N}$ and lidar-to-camera $\textbf{A}_{l2c}^{T} \in \mathbb{R}^{N\times N}$.
For student model, we adopt the same operates as the teacher model, we can generate  $\textbf{F}_{C_{sub1}}^{S} \in \mathbb{R}^{N\times  (P^2C)}$ and  $\textbf{F}_{C_{subs}}^{S} \in \mathbb{R}^{N\times  (P^2C)}$.  
%%%%%%%%%%%%%%%%%%%%%%%%%%%
The cross-modal attention distribution of the student $\textbf{A}_{c2l}^{S}$, $\textbf{A}_{l2c}^{S}$are computed as follows:
\begin{equation}
\label{eq3}
\begin{array}{l}
A_{c 2 l}^{S}=\operatorname{softmax}\left(\frac{F_{C_{sub1}}^{S} \cdot F_{C_{sub2}}^{S \top}}{\sqrt{D_{k}}}\right)
\end{array}
\end{equation}
\begin{equation}
\label{eq4}
\begin{array}{l}
A_{l 2 c}^{S}=\operatorname{softmax}\left(\frac{F_{C_{sub2}}^{S} \cdot F_{C_{sub1}}^{S \top}}{\sqrt{D_{k}}}\right)
\end{array}
\end{equation}
where $F_{C_{sub1}}$ and  $F_{C_{sub2}}$ are different subspace BEV features in the student model.
\fi

To this end, we propose the cross-modal relation distillation and employ a KL-divergence loss to align the cross-modal attention $\textbf{A}_{c2l}^{S}$ and $\textbf{A}_{l2c}^{S}$ of the student with $\textbf{A}_{c2l}^{T}$ and $\textbf{A}_{l2c}^{T}$ of the teacher model:
\begin{equation}
\label{eq5}
\begin{array}{l}
\mathcal{L}_{relation} = D_{KL}(A_{c2l}^{T}   || A_{c2l}^{S}) + D_{KL}(A_{l2c}^{T}  ||  A_{l2c}^{S}).
\end{array}
\end{equation}
\subsubsection{Dual-level Feature Distillation.}
To facilitate the student model to absorb the rich semantic/geometric knowledge from the teacher model, we take advantage of the fused BEV features for the feature-level distillation.
Specifically, we leverage the low-level fused BEV feature of the teacher $\textbf{F}_{fused}^{T}$ as the supervisory signal for learning the counterpart of the student $\textbf{F}_{fused}^{S}$ via an MSE loss, \textit{i.e.},
\begin{equation}
\label{eq6}
\begin{array}{l}
\mathcal{L}_{low} = \operatorname{MSE} (\textbf{F}_{fused}^{T},\textbf{F}_{fused}^{S}).
\end{array}
\end{equation}
In addition, we further propose the high-level feature distillation $\mathcal{L}_{high}$ to align $\textbf{F}_{high}^{T}$ and $\textbf{F}_{high}^{S}$, which are generated by the Map Encoder. $\mathcal{L}_{high}$ is defined as:
\begin{equation}
\label{eq7}
\begin{array}{l}
\mathcal{L}_{high} = \operatorname{MSE} (\textbf{F}_{high}^{T},\textbf{F}_{high}^{S}).
\end{array}
\end{equation}
Formally, the dual-level feature distillation loss $\mathcal{L}_{features}$ is the sum of low-level distillation loss $\mathcal{L}_{low}$ and high-level distillation loss $\mathcal{L}_{high}$, \textit{i.e.},
\begin{equation}
\label{eq8}
\begin{array}{l}
\mathcal{L}_{feature} = \mathcal{L}_{low} + \mathcal{L}_{high}.
\end{array}
\end{equation}
We use $\mathcal{L}_{feature}$ as one of the distillation objectives to enable the student model to benefit from the teacher model implicitly during training.

\subsubsection{Map Head  Distillation.}
After the Map Encoder, the high-level BEV feature in the student model is fed into the HD Map Head to produce the prediction in the same way as the teacher model. To make the final prediction of the student close to that of the teacher, we further propose the map head distillation. Specifically, we use the predictions generated by the teacher model as pseudo labels to supervise the student model via the $\mathcal{L}_{head}$ loss.
To achieve the goal, we need to construct the correspondence between the predictions of the student and the teacher. 
Suppose the classification and point predictions from the teacher model are $\textbf{F}_{cls}^{T}$ and $\textbf{F}_{point}^{T}$ respectively, and those from the student can be represented as $\textbf{F}_{cls}^{S}$ and $\textbf{F}_{point}^{S}$ respectively. The $\mathcal{L}_{head}$ loss consists of two parts, \textit{i.e.}, the classification loss $\mathcal{L}_{cls}$ for map elements classification and the point2point loss $\mathcal{L}_{point}$ for point position regression:
\begin{equation}
\label{eq9}
\begin{aligned}
  \mathcal{L}_{head} &= \mathcal{L}_{cls} + \mathcal{L}_{point}\\
  &=\mathcal{L}_{Focal}(\textbf{F}_{cls}^{T},\textbf{F}_{cls}^{S}) + \mathcal{L}_{p2p}(\textbf{F}_{point}^{T},\textbf{F}_{point}^{S}),
\end{aligned}
\end{equation}
where $\mathcal{L}_{Focal}$ denotes the Focal loss~\cite{mukhoti2020calibrating} and $\mathcal{L}_{p2p}$ denotes the Manhattan distance~\cite{malkauthekar2013analysis} between $\textbf{F}_{point}^{T}$ and $\textbf{F}_{point}^{S}$.

\subsection{Overall Training}
\label{sec3.3}
To facilitate knowledge transfer from the multi-modal fusion-based teacher model to the camera-based student model, we integrate the map loss $\mathcal{L}_{map}$ with the above distillation losses, including the cross-modal relation distillation loss ($\mathcal{L}_{relation}$), the dual-level feature distillation loss ($\mathcal{L}_{feature}$), and the map head distillation loss ($\mathcal{L}_{head}$).
The overall training objective can be formulated as:
\begin{equation}
	\label{eq10}
	\begin{aligned}	&\mathcal{L}=\mathcal{L}_{map}+\lambda_{1}\mathcal{L}_{relation}+\lambda_{2}\mathcal{L}_{feature} +\lambda_{3}\mathcal{L}_{head},
	\end{aligned}
\end{equation}
where $\lambda_{1}$, $\lambda_{2}$ and $\lambda_{3}$ are hyper-parameters for balancing these terms. The map loss $\mathcal{L}_{map}$ is calculated following~\cite{MapTR}, which is composed of three parts, \textit{i.e.}, classification loss, point2point loss, and edge direction loss.

% --------------------------------------------------------------------------
\section{Experiments}
\label{sec4}
\subsection{Experimental Settings}
\textbf{Datasets.} We evaluate our method on the widely-used challenging nuScenes~\cite{20cvprnuscense} dataset following the standard setting of previous methods~\cite{MapTR,li2022hdmapnet,liu2023vectormapnet}. 
The nuScenes dataset contains 1,000 sequences of recordings collected by autonomous driving cars. 
Each sample is annotated at 2Hz and contains 6 camera images covering $360^\circ$ horizontal FOV of the ego-vehicle.
Following~\cite{MapTR,li2022hdmapnet,liu2023vectormapnet}, 
three kinds of map elements are chosen for fair evaluation – pedestrian crossing, lane divider, and road boundary.

\textbf{Evaluation Metrics.}
We adopt the evaluation metrics used in previous works~\cite{MapTR,li2022hdmapnet,liu2023vectormapnet}. Specifically, average precision (AP) is used to evaluate the map construction quality.
Chamfer distance $D_{Chamfer}$ is
used to determine whether the prediction and GT are matched or not. We calculate the $AP_\tau$ under
several $D_{Chamfer}$ thresholds ($\tau \in T =\{0.5,1.0,1.5\} $), and then average across all thresholds as the final mean AP (\textit{mAP}) metric:
\begin{equation}
\label{eq11}
\begin{array}{l}
mAP\ =\ \frac{1}{\left|T\right|}\sum\limits_{\tau\in T}{AP}_\tau.
\end{array}
\end{equation}
The perception ranges are [-15.0m, 15.0m]/[-30.0m, 30.0m] for X/Y-axes.

\textbf{Model and Training Details.}
MapDistill is trained with 8 NVIDIA RTX A6000 GPUs. 
For the teacher model, we first establish a baseline method of fusion-based  HD map construction based on MapTR~\cite{MapTR}. 
The fused MapTR model uses ResNet50~\cite{2016cvprresnet} and SECOND~\cite{2018seoncd} as the backbone and employ GKT~\cite{2022GKT} as the default 2D-to-BEV module. For the student model, we adopt MapTR's camera branch as the base, and introduce the dual BEV transform module to facilitate cross-modal knowledge distillation. 
Note that, the student model adopts ResNet18~\cite{2016cvprresnet} as the backbone. 
Moreover, we adopt the AdamW optimizer~\cite{2019iclradamw} for all our experiments.
The setting of hyper-parameters $\lambda_1$, $\lambda_2$, and $\lambda_3$ is discussed extensively in the ablation studies.
We set the mini-batch size to 64, and use a step-decayed learning rate with an initial value of $4e^{-3}$.

\begin{table*}[t]
\begin{center}
\setlength{\belowcaptionskip}{-0.0001cm}
\caption{
Performance analysis of MapDistill on nuScenes val set.
``L'' and ``C'' represent the LiDAR and camera, respectively.
``Effi-B0'', ``R18'', ``R50'', and ``Sec'' are short for EfficientNet-B0~\cite{tan2019efficientnet}, ResNet18~\cite{2016cvprresnet}, ResNet50~\cite{2016cvprresnet}, and SECOND~\cite{2018seoncd}, respectively.
We adopt the MapTR method to build the teacher model and the student model. Note that the directly-trained MapTR models in the red region are selected as teachers.
Our proposed MapDistill outperforms all existing approaches in both single-class APs and the overall mAP by a significant margin.
$\dag$ denotes our re-implementation following the setting in the paper. Best viewed in color.
}
\scalebox{0.75}{
  \begin{tabular}{p{2.8cm}|p{1.8cm}p{2cm}|p{2cm}p{1.4cm}|p{1.4cm}p{1.4cm}p{1.4cm}p{1.4cm}}
  
  \hline
 \rowcolor{black!10} \makecell[l]{Method}&  \makecell[c]{Student\\ Modality} &  \makecell[c]{Teacher\\ Modality}& \makecell[c]{Backbone} & \makecell[c]{Epochs}& \makecell[c]{AP$_{ped.}$} &\makecell[c]{AP$_{div.}$}&\makecell[c]{AP$_{bou.}$}&\makecell[l]{mAP}\\
  \midrule
\makecell[l]{HDMapNet~\cite{li2022hdmapnet}}& \makecell[c]{C}& \makecell[c]{$-$}& \makecell[c]{Effi$-$B0} & \makecell[c]{30}& \makecell[c]{14.4} &\makecell[c]{21.7}&\makecell[c]{33.0}&\makecell[l]{23.0}\\

\makecell[l]{VectorMapNet~\cite{liu2023vectormapnet}}& \makecell[c]{C}& \makecell[c]{$-$}& \makecell[c]{R50} & \makecell[c]{110}& \makecell[c]{36.1} &\makecell[c]{47.3}&\makecell[c]{39.3}&\makecell[l]{40.9}\\

\makecell[l]{MapVR~\cite{zhang2023online}}& \makecell[c]{C}& \makecell[c]{$-$}& \makecell[c]{R50} & \makecell[c]{24}& \makecell[c]{47.7} &\makecell[c]{54.4}&\makecell[c]{51.4}&\makecell[l]{51.2}\\
   
\makecell[l]{PivotNet~\cite{ding2023pivotnet}}& \makecell[c]{C}& \makecell[c]{$-$}& \makecell[c]{R50} & \makecell[c]{30}&\makecell[c]{58.5} &\makecell[c]{53.8}&\makecell[c]{59.6}&\makecell[l]{57.4}\\
       
\makecell[l]{BeMapNet~\cite{qiao2023end}}& \makecell[c]{C}& \makecell[c]{$-$}& \makecell[c]{R50} & \makecell[c]{30}& \makecell[c]{62.3} &\makecell[c]{57.7}&\makecell[c]{59.4}&\makecell[l]{59.8}\\

\rowcolor{red!10} \makecell[l]{MapTR~\cite{MapTR}}& \makecell[c]{C}& \makecell[c]{$-$}& \makecell[c]{R50} & \makecell[c]{24}& \makecell[c]{45.3} &\makecell[c]{51.5}&\makecell[c]{53.1}&\makecell[l]{50.3}\\

\rowcolor{red!10} \makecell[l]{MapTR~\cite{MapTR}}&  \makecell[c]{L}& \makecell[c]{$-$}& \makecell[c]{Sec} & \makecell[c]{24}& \makecell[c]{48.5} &\makecell[c]{53.7}&\makecell[c]{64.7}&\makecell[l]{55.6}\\

\rowcolor{red!10}  \makecell[l]{MapTR~\cite{MapTR}}&  \makecell[c]{C $\&$ L}& \makecell[c]{$-$}& \makecell[c]{R50 $\&$ Sec} & \makecell[c]{24}& \makecell[c]{55.9} &\makecell[c]{62.3}&\makecell[c]{69.3}&\makecell[l]{62.5}\\

\hline
\rowcolor{green!10} \makecell[l]{MapTR~\cite{MapTR}}&  \makecell[c]{C}& \makecell[c]{$-$}& \makecell[c]{R18} & \makecell[c]{110}& \makecell[c]{39.6} &\makecell[c]{49.9}&\makecell[c]{48.2}&\makecell[l]{45.9}\\
 
\hline
\makecell[l]{BEV-LGKD{$\dag$}~\cite{li2022bev}}&  \makecell[c]{C}& \makecell[c]{C} & \makecell[c]{R18} & \makecell[c]{110}& \makecell[c]{42.2} &\makecell[c]{47.6}&\makecell[c]{49.7}
&\makecell[l]{46.5$_{+0.6}$}\\

\makecell[l]{BEVDistill{$\dag$}~\cite{chen2022bevdistill}}&  \makecell[c]{C}& \makecell[c]{L}& \makecell[c]{R18} & \makecell[c]{110}& \makecell[c]{42.4} &\makecell[c]{48.5}&\makecell[c]{50.2}&\makecell[l]{47.1$_{+1.2}$}\\

\makecell[l]{UniDistill{$\dag$}~\cite{zhou2023unidistill}}& \makecell[c]{C}& \makecell[c]{C$\&$L}& \makecell[c]{R18} & \makecell[c]{110}& \makecell[c]{43.9} &\makecell[c]{48.6}&\makecell[c]{52.1}&\makecell[l]{48.2$_{+2.3}$}\\
   
\rowcolor{blue!10} \makecell[l]{MapDistill}& \makecell[c]{C}& \makecell[c]{C}& \makecell[c]{R18} &\makecell[c]{110}& \makecell[c]{43.3} &\makecell[c]{48.8}&\makecell[c]{51.9}&\makecell[l]{48.0$_{+2.1}$}\\

\rowcolor{blue!10} \makecell[l]{MapDistill}& \makecell[c]{C}& \makecell[c]{L}& \makecell[c]{R18} &\makecell[c]{110}& \makecell[c]{45.9} &\makecell[c]{50.7}&\makecell[c]{53.6}&\makecell[l]{50.1$_{+4.2}$}\\

\rowcolor{blue!10} \makecell[l]{\textbf{MapDistill}}& \makecell[c]{\textbf{C}}& \makecell[c]{\textbf{C $\&$ L}}& \makecell[c]{\textbf{R18}} &\makecell[c]{\textbf{110}}& \makecell[c]{\textbf{49.2}} &\makecell[c]{\textbf{54.5}}&\makecell[c]{\textbf{57.1}}&\makecell[l]{\textbf{53.6}$_{+7.7}$}\\
  \hline
  \bottomrule
  \end{tabular}}
   \label{tab1}
\end{center}
 \vspace{-2.5em}
\end{table*}

\subsection{Comparison with the State-of-the-Arts}
We compare our method with several state-of-the-art baselines across two categories, \textit{i.e.}, camera-based HD map construction methods, and customized KD methods which were originally designed for BEV-based 3D object detection.
For camera-based HD map construction methods, we directly report the results from the corresponding papers. 
For KD-based methods, we implement three methods for BEV-based 3D object detection and modify them for the HD map construction task, including BEV-LGKD~\cite{li2022bev}, BEVDistill~\cite{chen2022bevdistill}, and UnDistill~\cite{zhou2023unidistill}.
For fairness, we use the same teacher and student models as our method.

Tab.~\ref{tab1} shows that:
(1) KD methods originally designed for BEV-based 3D object detection fail to achieve satisfying results due to task discrepancies between 3D object detection and HD map construction.
(2) Intra-modal distillation between camera-only teacher and student models cannot learn accurate 3D information due to the limited capacity of the teacher model for inferring 3D geometry, and the gain is only 0.6 mAP by BEV-LGKD and 2.1 mAP by our MapDistill.
(3) Cross-modal distillation between the LiDAR teacher and the camera student enables learning useful 3D information from the teacher but suffers from the large cross-modal gap, achieving the improved gain of 1.2 mAP by BEVDistill and 4.2 mAP by our MapDistill.
(4) Our MapDistill with the fusion-based teacher enables effective knowledge distillation within/between modalities while enjoying cost-effective camera-only deployment, achieving the most significant gain of 7.7 mAP and surpassing UniDistill by 5.4 mAP.

\subsection{Ablation Study}
\textbf{Effect of $\mathcal{L}_{relation}$, $\mathcal{L}_{feature}$, and $\mathcal{L}_{head}$.}
We conduct an ablation study on the components in MapDistill and summarize our results in Tab.~\ref{tab2}.
We evaluate model variants using different combinations of the proposed distillation losses, including $\mathcal{L}_{relation}$, $\mathcal{L}_{feature}$, and $\mathcal{L}_{head}$.

We first investigate the effect of each distillation loss function.
In model variants (a), (b), and (c), we use different distillation losses to distill the student model separately.
The experimental results show that all model variants get improved performance compared to the baseline model, verifying the effectiveness of the proposed distillation losses.
Moreover, the results of model variants (d), (e), and (f) prove that different distillation losses are complementary to each other.
Finally, using all the proposed distillation losses together, we arrive at the full MapDistill method, which achieves the overall best performance of 53.6 mAP, significantly surpassing the baseline's performance of 45.9 mAP. 
The ablation study results show that each of the distillation losses in MapDistll provides a meaningful contribution to improving the student model performance.
Notably, these losses are only calculated during training, which brings no computational overhead during inference.

\begin{table}[t]
\begin{center}
\caption{Ablation study on the components in MapDistill. 
}
\scalebox{0.85}{
  \begin{tabular}{p{1.3cm}p{1.4cm}p{1.4cm}p{1.4cm}|p{1cm}p{1cm}p{1cm}p{0.8cm}}
  
  \hline
 \rowcolor{black!10} \makecell[c]{Setting}&\makecell[c]{$\mathcal{L}_{relation}$}& \makecell[c]{$\mathcal{L}_{feature}$}& \makecell[c]{$\mathcal{L}_{head}$} & \makecell[c]{AP$_{ped.}$} &\makecell[c]{AP$_{div.}$}&\makecell[c]{AP$_{bou.}$}&\makecell[c]{mAP}\\
  \midrule
  
   \makecell[c]{Baseline}&\makecell[c]{\XSolidBrush}& \makecell[c]{\XSolidBrush}& \makecell[c]{\XSolidBrush}& \makecell[c]{39.6} &\makecell[c]{49.9}&\makecell[c]{48.2}&\makecell[c]{45.9}\\
    \hline
\makecell[c]{a}& \makecell[c]{\CheckmarkBold}& \makecell[c]{\XSolidBrush}& \makecell[c]{\XSolidBrush}& \makecell[c]{44.1} &\makecell[c]{49.7}&\makecell[c]{52.4}&\makecell[c]{48.8}\\
   \makecell[c]{b}& \makecell[c]{\XSolidBrush}& \makecell[c]{\CheckmarkBold}& \makecell[c]{\XSolidBrush}& \makecell[c]{44.3} &\makecell[c]{49.4}&\makecell[c]{51.5}&\makecell[c]{48.4}\\
   \makecell[c]{c}&\makecell[c]{\XSolidBrush}& \makecell[c]{\XSolidBrush}& \makecell[c]{\CheckmarkBold}& \makecell[c]{44.2} &\makecell[c]{50.1}&\makecell[c]{52.7}&\makecell[c]{49.0}\\
   \hline
 \makecell[c]{d}&\makecell[c]{\CheckmarkBold}& \makecell[c]{\CheckmarkBold}& \makecell[c]{\XSolidBrush}& \makecell[c]{45.4} &\makecell[c]{51.4}&\makecell[c]{54.1}&\makecell[c]{50.3}\\
   \makecell[c]{e}& \makecell[c]{\XSolidBrush}& \makecell[c]{\CheckmarkBold}& \makecell[c]{\CheckmarkBold}& \makecell[c]{46.3} &\makecell[c]{51.8}&\makecell[c]{54.3}&\makecell[c]{50.8}\\
    \makecell[c]{f}& \makecell[c]{\CheckmarkBold}& \makecell[c]{\XSolidBrush}& \makecell[c]{\CheckmarkBold}& \makecell[c]{46.5} &\makecell[c]{52.3}&\makecell[c]{54.5}&\makecell[c]{51.1}\\
     \hline
   \rowcolor{blue!10}\makecell[c]{g}&\makecell[c]{\CheckmarkBold}& \makecell[c]{\CheckmarkBold}& \makecell[c]{\CheckmarkBold}& \makecell[c]{\textbf{49.2}} &\makecell[c]{\textbf{54.5}}&\makecell[c]{\textbf{57.1}}&\makecell[c]{\textbf{53.6}}\\
  \bottomrule
    \label{tab2}
  \end{tabular}}
\end{center}
 \vspace{-4.6em}
\end{table}

\textbf{Ablations on the cross-modal relation distillation.}
We investigate the choice of relation distillation loss in our method. 
The ablation variants include training without relation distillation loss (MapDistill (w/o $\mathcal{L}_{relation}$)), uni-modal relation distillation (Uni-modal Rel.), cross-modal relation distillation (Cross-modal Rel.), and the hybrid relation distillation (hybrid Cross-modal and Uni-modal). Note that uni-modal relation distillation means replacing the cross-modal attention matrices $\textbf{A}_{c2l}^{S/T}$ and $\textbf{A}_{l2c}^{S/T}$ in Eq.~\ref{eq5} with the uni-modal ones $\textbf{A}_{c2c}^{S/T}$ and $\textbf{A}_{l2l}^{S/T}$.
We explore which relation (cross-modal or uni-modal) is more critical.
As shown in Tab.~\ref{tab-cross}, employing cross-modal relation distillation achieves more improvements.
Furthermore, we find that using only cross-modal relation for distillation performs better than using both cross-modal and uni-modal relations.
These observations validate that cross-modal interactions encode useful knowledge and can be transferred to the student model for improving HD Map construction.

\textbf{Ablations on the dual-level feature distillation.}
To explore the impact of BEV feature distillation at different levels, we train the model by using low-level or high-level feature distillation solely and present the results in Tab.~\ref{tab-dual}.
We design the following model variants:
(1) MapDistill (w/o $\mathcal{L}_{feature}$): we remove the feature distillation loss from MapDistill.
(2) Low-level (only) and High-level (only) mean that the MapDistill model is trained only using low-level BEV feature distillation or high-level BEV feature distillation, respectively.
(3) Dual-level (ours): we use dual-level feature distillation (the default setting in our MapDistill) to train the model.
The results of Low-level (only) and High-level (only) are inferior to the Dual-level (ours), verifying the effectiveness of distilling both low-level and high-level BEV features simultaneously.

\textbf{Ablations on the map head distillation.}
In this ablation, we conduct detailed experiments on the loss selection for both map elements classification and point position regression. We design the following model variants:
(1) MapDistill (w/o $\mathcal{L}_{head}$): we train the model without the map head distillation loss; 
(2) $\mathcal{L}_{head}$ (w/o $\mathcal{L}_{point}$): we remove the point2point loss from the map head distillation loss; 
(3) $\mathcal{L}_{head}$ (w/o $\mathcal{L}_{cls}$): we remove the classification loss from the map head distillation loss; 
(4) Using both $\mathcal{L}_{cls}$ and $\mathcal{L}_{point}$ (the default setting in our MapDistill). 
As shown in Tab.~\ref{tab-map}, the results of $\mathcal{L}_{head}$ (w/o $\mathcal{L}_{cls}$) and $\mathcal{L}_{head}$ (w/o $\mathcal{L}_{point}$) are inferior to the default setting, verifying the effectiveness of transferring knowledge of both map elements categories and point positions from the teacher to the student.

\begin{table*}[t]
\centering
\caption{Ablation experiments to validate our distillation losses. }
\vspace{-15pt}
\label{tab3}
~~
\begin{subtable}[t]{0.45\linewidth}
\centering\small
\addtolength{\tabcolsep}{1pt} 
\caption{Cross-modal relation distillation loss}
\vspace{-5pt}
\scalebox{0.7}{
 \begin{tabular}{p{4.0cm}|p{1cm}p{1cm}p{1cm}p{0.8cm}}
  \hline
 \rowcolor{black!10}  \makecell[c]{Method} & \makecell[c]{AP$_{ped.}$} &\makecell[c]{AP$_{div.}$}&\makecell[c]{AP$_{bou.}$}&\makecell[c]{mAP}\\
  \midrule
  \makecell[l]{MapDistill (w/o $\mathcal{L}_{relation}$)}&\makecell[c]{46.3} &\makecell[c]{51.8}&\makecell[c]{54.3}&\makecell[c]{50.8}\\
    \makecell[l]{+Uni-modal Relation}& \makecell[c]{48.0} &\makecell[c]{52.9}&\makecell[c]{55.1}&\makecell[c]{52.0}\\
 \makecell[l]{+Hybrid Relation}&  \makecell[c]{48.3} &\makecell[c]{53.4}&\makecell[c]{55.5}&\makecell[c]{52.4}\\
  \hline
     \rowcolor{blue!10} \makecell[l]{\textbf{+Cross-modal Relation}}& \makecell[c]{\textbf{49.2}} &\makecell[c]{\textbf{54.5}}&\makecell[c]{\textbf{57.1}}&\makecell[c]{\textbf{53.6}}\\
  \bottomrule
  \end{tabular}
}
\label{tab-cross}
\end{subtable}
~

\begin{subtable}[t]{0.45\linewidth}
\centering\small
\addtolength{\tabcolsep}{1pt} 
\vspace{3pt}
\caption{Dual-level feature distillation loss}
\vspace{-5pt}
\scalebox{0.68}{
  \begin{tabular}{p{3.8cm}|p{1cm}p{1cm}p{1cm}p{0.8cm}}
  \hline
 \rowcolor{black!10}  \makecell[c]{Method} & \makecell[c]{AP$_{ped.}$} &\makecell[c]{AP$_{div.}$}&\makecell[c]{AP$_{bou.}$}&\makecell[c]{mAP}\\
  \midrule
  \makecell[l]{MapDistill (w/o $\mathcal{L}_{feature}$)}&\makecell[c]{46.5} &\makecell[c]{52.3}&\makecell[c]{54.5}&\makecell[c]{51.1}\\
    \makecell[l]{+Low-level (only)}& \makecell[c]{48.4} &\makecell[c]{53.7}&\makecell[c]{56.0}&\makecell[c]{52.7}\\
 \makecell[l]{+High-level (only)}&  \makecell[c]{48.7} &\makecell[c]{53.9}&\makecell[c]{56.1}&\makecell[c]{52.9}\\
  \hline
     \rowcolor{blue!10} \makecell[l]{\textbf{+Dual-level (ours)}}& \makecell[c]{\textbf{49.2}} &\makecell[c]{\textbf{54.5}}&\makecell[c]{\textbf{57.1}}&\makecell[c]{\textbf{53.6}}\\
  \bottomrule
  \end{tabular}
}
\label{tab-dual}
\end{subtable}
~~
\begin{subtable}[t]{0.45\linewidth}
\centering\small
\setlength{\tabcolsep}{6.2pt}
\vspace{3pt}
\caption{Map head distillation loss}
\vspace{-5pt}
\scalebox{0.64}{
 \begin{tabular}{p{1.2cm}p{1.2cm}|p{1cm}p{1cm}p{1cm}p{0.8cm}}
  \hline
 \rowcolor{black!10}  \makecell[c]{$\mathcal{L}_{cls}$}& \makecell[c]{$\mathcal{L}_{point}$} & \makecell[c]{AP$_{ped.}$} &\makecell[c]{AP$_{div.}$}&\makecell[c]{AP$_{bou.}$}&\makecell[c]{mAP}\\
  \midrule
    \makecell[c]{\XSolidBrush}& \makecell[c]{\XSolidBrush}& \makecell[c]{45.4} &\makecell[c]{51.4}&\makecell[c]{54.1}&\makecell[c]{50.3}\\
 \makecell[c]{\CheckmarkBold}& \makecell[c]{\XSolidBrush}& \makecell[c]{47.3} &\makecell[c]{52.8}&\makecell[c]{55.3}&\makecell[c]{51.8}\\
 \makecell[c]{\XSolidBrush}& \makecell[c]{\CheckmarkBold}& \makecell[c]{47.1} &\makecell[c]{53.0}&\makecell[c]{55.6}&\makecell[c]{51.9}\\
 \hline
     \rowcolor{blue!10} \makecell[c]{\CheckmarkBold}& \makecell[c]{\CheckmarkBold}& \makecell[c]{\textbf{49.2}} &\makecell[c]{\textbf{54.5}}&\makecell[c]{\textbf{57.1}}&\makecell[c]{\textbf{53.6}}\\
  \bottomrule
  \end{tabular}
}
\label{tab-map}
\end{subtable}
~\vspace{-15pt}
\end{table*}

\makeatletter\def\@captype{table}\makeatother
\begin{minipage}{.58\textwidth}
\scriptsize
\centering
\caption{Ablation study of Dual BEV Transform Module.}
\begin{tabular}[t]{ccc|cccc}  
  \hline

\rowcolor{black!10} &  \makecell[c]{subspace1}& \makecell[c]{subspace2} & \makecell[c]{AP$_{ped.}$} &\makecell[c]{AP$_{div.}$}&\makecell[c]{AP$_{bou.}$}&\makecell[c]{mAP}\\
  \midrule

   \makecell[l]{(a)}& \makecell[c]{GKT}& \makecell[c]{\XSolidBrush}& \makecell[c]{44.9} &\makecell[c]{49.6}&\makecell[c]{52.8}&\makecell[c]{49.1}\\
    \hline
   \multirow{3}{*}{\makecell[l]{(b)}}& \makecell[c]{LSS}& \makecell[c]{LSS}& \makecell[c]{45.9} &\makecell[c]{51.2}&\makecell[c]{54.4}&\makecell[c]{50.5}\\
   &   \makecell[c]{GKT}& \makecell[c]{GKT}& \makecell[c]{46.7} &\makecell[c]{51.6}&\makecell[c]{54.5}&\makecell[c]{50.9}\\
  &   \makecell[c]{Deform.}& \makecell[c]{Deform.}& \makecell[c]{46.8} &\makecell[c]{51.6}&\makecell[c]{54.6}&\makecell[c]{51.0}\\
    \hline
    %%%%%%第1种组合
 \multirow{5}{*}{\makecell[l]{(c)}} &   \makecell[c]{GKT}& \makecell[c]{Deform.}& \makecell[c]{47.1} &\makecell[c]{53.2}&\makecell[c]{56.2}&\makecell[c]{52.1}\\
   &  \makecell[c]{Deform.}& \makecell[c]{GKT}& \makecell[c]{47.3} &\makecell[c]{53.4}&\makecell[c]{56.1}&\makecell[c]{52.3}\\
      %%%%%%第2种组合
   & \makecell[c]{LSS}& \makecell[c]{Deform.}& \makecell[c]{48.9} &\makecell[c]{53.9}&\makecell[c]{56.2}&\makecell[c]{53.0}\\
 & \makecell[c]{Deform.}& \makecell[c]{LSS}& \makecell[c]{48.7} &\makecell[c]{53.8}&\makecell[c]{55.9}&\makecell[c]{52.8}\\
    
%%%%%%第3种组合
&\makecell[c]{LSS}& \makecell[c]{GKT}& \makecell[c]{49.1} &\makecell[c]{54.2}&\makecell[c]{56.7}&\makecell[c]{53.3}\\
 \rowcolor{blue!10} &  \textbf{GKT}& \textbf{LSS}& \textbf{49.2} &\textbf{54.5}&\textbf{57.1}&\textbf{53.6}\\
  \bottomrule
    \label{tab7}
  \end{tabular}

\label{tab:710}
\end{minipage}
\hfill
\makeatletter\def\@captype{figure}\makeatother
\begin{minipage}{.35\textwidth}
\centering 
\includegraphics[height=3.4cm]{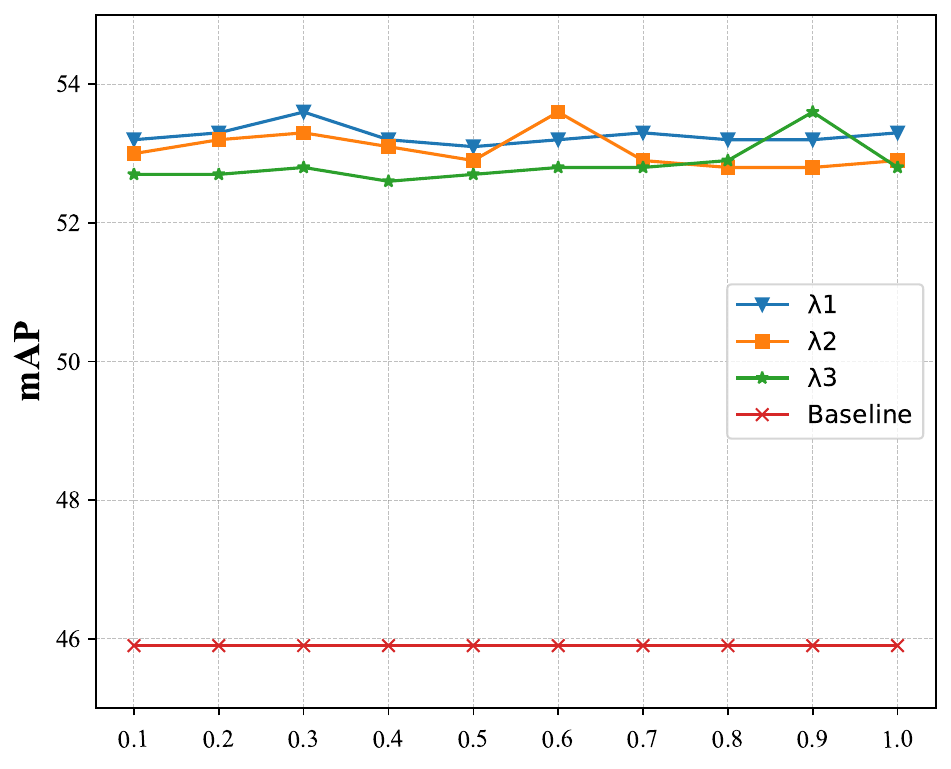}
\caption{Sensitivity of hyper-parameters.}
\label{fig4}
\end{minipage}

\textbf{Ablation study of the Dual BEV Transform Module.} 
We further conduct ablation studies on the design choice of the Dual BEV Transform Module, \textit{i.e.}, using different 2D-to-BEV methods to obtain subspace BEV features, to verify which combination performs most effectively. We choose three state-of-the-art 2D-to-BEV methods in this study, including LSS~\cite{philion2020lift}, Deformable Attention~\cite{22eccvbevformer} and GKT~\cite{2022GKT}.
The experiment consists of two groups, 
(1) both subspaces use the same 2D-to-BEV method, and (2) the two subspaces use different 2D-to-BEV methods.
As shown in Tab.~\ref{tab7},  the experimental results reveal some interesting findings:
(1) Using the same 2D-to-BEV method only slightly outperforms the single-branch baseline in (a), implying that using the homogeneous BEV feature space makes it difficult to imitate the cross-modal interactions in the teacher model.
(2) Using different 2D-to-BEV methods consistently outperforms the baseline and all model variants in (b). It is reasonable since the cross-modal relation of the teacher is calculated based on BEV features from different modalities, using heterogeneous BEV feature spaces makes it possible to learn distinct BEV features and thus could imitate the cross-modal interactions.
Specifically, the combination of LSS and GKT achieves the best results.
These observations validate the motivation for devising dual BEV spaces using different 2D-to-BEV methods.

\textbf{Ablation study of various HD map construction methods.}
To explore the compatibility of MapDistill with different HD map construction methods, we comprehensively investigate two popular methods and show the results in Tab.~\ref{tab-teacher}.
Specifically,  Teacher model-1 and Teacher model-2 mean the MapTR variant model whose camera branch uses Swin-Tiny backbone to extract image features and the most advanced MapTRv2 (improving MapTR with both network design and training strategy techniques), respectively. 
Note that, both student models employ Resnet18 as the backbone to extract the multi-view features.
The experimental results demonstrate that ``Great teachers spawn exceptional students''. As the proficient teacher model has acquired valuable knowledge for HD map construction, the student model can effectively
leverage this knowledge through KD techniques (\textit{e.g.}, the proposed MapDistill), enhancing its ability to perform the same task. Moreover, the results of consistent performance improvements show that our method is effective with different teacher models.

\textbf{Ablation study of various student models.}
To explore the generalization capability of MapDistill with different student models, we comprehensively investigate two popular backbone networks as the backbone of the student model and show the results in Tab.~\ref{tab-student}.
Specifically,  Student model-\uppercase\expandafter{\romannumeral1} and Student model-\uppercase\expandafter{\romannumeral2} mean that 
the student model employs Resnet50 and Swin-Tiny as the backbone to extract the multi-view features, respectively.
And here we use MapTR-Teacher, which is the R50\&Sec fusion model in Tab.~\ref{tab1}, as the teacher model.
Experimental results show that our method consistently achieves excellent results, proving the effectiveness and generalization ability of our method.

\begin{table*}[t]
\centering
\caption{\textbf{
Ablation experiments to verify the generalization ability of MapDistill.}}
\vspace{-15pt}
\label{tab5}
\begin{subtable}[t]{0.45\linewidth}
\centering\small
\addtolength{\tabcolsep}{1pt} 
\vspace{3pt}
\caption{Different HD map construction methods}
\vspace{-5pt}
\scalebox{0.56}{
   \begin{tabular}{p{2.5cm}|p{1.8cm}|p{1cm}p{1cm}p{1cm}p{1.4cm}}
  \hline
 \rowcolor{black!10}  \makecell[c]{Method} & \makecell[c]{Backbone} &\makecell[c]{AP$_{ped.}$} &\makecell[c]{AP$_{div.}$}&\makecell[c]{AP$_{bou.}$}&\makecell[l]{mAP}\\
  \midrule
\makecell[c]{Teacher model-1}& \makecell[c]{SwinT$\&$Sec}& \makecell[c]{57.5} &\makecell[c]{63.3}&\makecell[c]{70.9}&\makecell[l]{63.9}\\
   
\makecell[c]{Student model-1}& \makecell[c]{R18}& \makecell[c]{39.6} &\makecell[c]{49.9}&\makecell[c]{48.2}&\makecell[l]{45.9}\\

\rowcolor{blue!10} \makecell[c]{\textbf{MapDistill}}& \makecell[c]{R18}&  \makecell[c]{\textbf{50.2}} &\makecell[c]{\textbf{55.5}}&\makecell[c]{\textbf{57.8}}&\makecell[l]{\textbf{54.5}$_{+8.6}$}\\

   \hline
\makecell[c]{Teacher model-2}& \makecell[c]{R50$\&$Sec}& \makecell[c]{65.6} &\makecell[c]{66.5}&\makecell[c]{74.8}&\makecell[l]{69.0}\\

\makecell[c]{Student model-2}& \makecell[c]{R18}& \makecell[c]{46.9} &\makecell[c]{55.1}&\makecell[c]{54.9}&\makecell[l]{52.3}\\

\rowcolor{blue!10} \makecell[c]{\textbf{MapDistill}}& \makecell[c]{R18}&  \makecell[c]{\textbf{53.9}} &\makecell[c]{\textbf{62.2}}&\makecell[c]{\textbf{61.5}}&\makecell[l]{\textbf{59.2}$_{+6.9}$}\\
  \bottomrule
  \end{tabular}
}
\label{tab-teacher}
\end{subtable}
~~
\begin{subtable}[t]{0.45\linewidth}
\centering\small
\setlength{\tabcolsep}{6.2pt}
\vspace{3pt}
\caption{Different student models}
\vspace{-5pt}
\scalebox{0.55}{

   \begin{tabular}{p{2.5cm}|p{1.4cm}|p{1cm}p{1cm}p{1cm}p{1.4cm}}
  \hline
 \rowcolor{black!10}  \makecell[c]{Method} & \makecell[c]{Backbone} & \makecell[c]{AP$_{ped.}$} &\makecell[c]{AP$_{div.}$}&\makecell[c]{AP$_{bou.}$}&\makecell[l]{mAP}\\
  \midrule
    \makecell[c]{MapTR-Teacher}& \makecell[c]{R50$\&$Sec}& \makecell[c]{55.9} &\makecell[c]{62.3}&\makecell[c]{69.3}&\makecell[l]{62.5}\\
     \makecell[c]{Student model-\uppercase\expandafter{\romannumeral1}}& \makecell[c]{R50}&\makecell[c]{46.3} &\makecell[c]{51.5}&\makecell[c]{53.1}&\makecell[l]{50.3}\\
   \rowcolor{blue!10} \makecell[c]{\textbf{MapDistill}}  &\makecell[c]{R50}& \makecell[c]{\textbf{51.3}} &\makecell[c]{\textbf{56.4}}&\makecell[c]{\textbf{57.6}}&\makecell[l]{\textbf{55.1}$_{+4.8}$}\\
   \hline
     \makecell[c]{MapTR-Teacher}&\makecell[c]{R50$\&$Sec}& \makecell[c]{55.9} &\makecell[c]{62.3}&\makecell[c]{69.3}&\makecell[l]{62.5}\\
     \makecell[c]{Student model-\uppercase\expandafter{\romannumeral2}}& \makecell[c]{Swin-T}&\makecell[c]{45.2} &\makecell[c]{52.7}&\makecell[c]{52.3}&\makecell[l]{50.1}\\
   \rowcolor{blue!10} \makecell[c]{\textbf{MapDistill}}  & \makecell[c]{Swin-T}&\makecell[c]{\textbf{50.1}} &\makecell[c]{\textbf{56.8}}&\makecell[c]{\textbf{58.9}}&\makecell[l]{\textbf{55.2}$_{+5.1}$}\\
  \bottomrule
  \end{tabular}
}
\label{tab-student}
\end{subtable}
~

\vspace{-25pt}
\end{table*}

\textbf{Sensitivity of hyper-parameters.}
We conduct experiments to investigate the impact of different hyper-parameter settings and report results on nuScenes val set, as shown in Fig.~\ref{fig4}. When one hyper-parameter is varied within a feasible range, the remaining hyper-parameters retain their default values: $\lambda_1=0.3$, $\lambda_2=0.6$, and $\lambda_3=0.9$. The results indicate that the performance remains relatively stable across a wide range (0.1 to 0.9) of values for $\lambda_1$, $\lambda_2$, and $\lambda_3$, suggesting its robustness to different hyper-parameters.
Note that, ``Baseline'' indicates the directly trained camera-based student model.

\begin{figure*}[!htbp]
	\centering
\setlength{\abovecaptionskip}{0.2em}
 	\includegraphics[width=.96\textwidth]{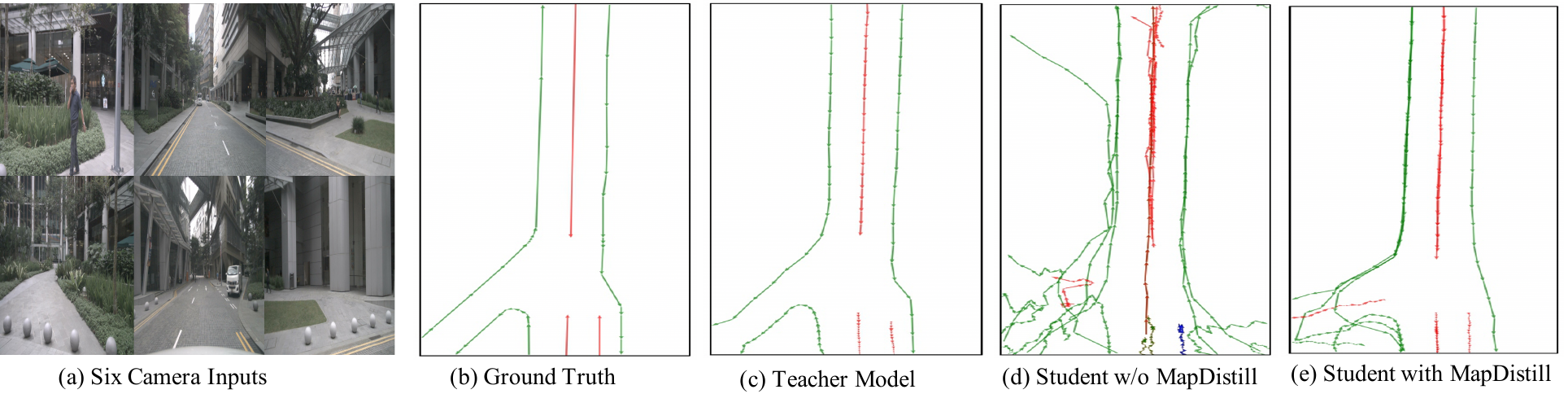}
	\caption{
 Qualitative results on nuScenes val set. (a) Six camera inputs. (b) Ground-truth vectorized HD map. (c) Result of the camera-LiDAR-based teacher model. (d) Result of the camera-based student model without MapDistill (Baseline). (e) Result of the camera-based student model with MapDistill. MapDistill helps correct substantial errors in the Baseline's predictions and improves its accuracy.
}
\label{fig5}
\vspace{-1.7em}
\end{figure*}

\subsection{Qualitative Results}
We present visualizations of vectorized HD map predictions to demonstrate the efficacy of MapDistill. 
As depicted in Fig.~\ref{fig5}, we compare predictions from various models, namely, the camera-LiDAR-based teacher model, the camera-based student model without MapDistill (referred to as ``Baseline''), and the camera-based student model with MapDistill. The mAP values of these models are 62.5, 45.9, and 53.6 respectively, as shown in Tab.~\ref{tab1}.
Note that a common threshold, which is set to 0.4, is employed to filter low-confidence map elements for visualizing the prediction results of all models.
We observe significant inaccuracies in the predictions made by the Baseline model. However, employing the MapDistill method substantially corrects these errors and enhances prediction accuracy.

%%%%%%%%%%%%%%%%%%%%%%%%%%%%%%%%%%%%%%%%%%%%%%%%%%%%%%%%%%%%%%%%%%%%%%%%%%%%%%%%
\section{Conclusion}
In this paper, we present a novel method called MapDistill for boosting efficient camera-based HD map construction via
camera-LiDAR fusion model distillation, yielding a cost-effective yet accurate solution. MapDistill is built upon a camera-LiDAR fusion teacher model, a lightweight camera-only student model, and a specifically designed Dual BEV Transform module. In addition, we present a comprehensive distillation scheme encompassing cross-modal relation distillation, dual-level feature distillation, and map head distillation, which facilitates knowledge transfer within and between different modalities and helps the student model achieve better performance. Extensive experiments and analysis validate the design choice and the effectiveness of our MapDistill. 
%In the future work, we will explore and apply the proposed approach to more camera-based perception tasks in BEV,  such as 3D object detection, segmentation and tracking.

\textbf{Limitations and Societal Impact.}
With the KD methodology, the student model may inherit the weakness of the teacher model. More specifically, if the teacher model is biased, or not robust to adverse weather conditions and/or long-tail scenarios, the student model is likely to behave similarly.
MapDistill enjoys cost-effective camera-only deployment, showing great potential in practical applications, such as autonomous driving.

\textbf{Acknowledgement.} This work was supported by the National Natural Science Foundation of China No.62106259 and the Beijing Natural Science Foundation under Grant L243008.

\clearpage
% ---- Bibliography ----
%
% BibTeX users should specify bibliography style 'splncs04'.
% References will then be sorted and formatted in the correct style.
%
\bibliographystyle{splncs04}
\bibliography{main}
\end{document}